\documentclass{vgtc}                          
\ifpdf
  \pdfoutput=1\relax                   
  \usepackage{graphicx}                
  \DeclareGraphicsExtensions{.pdf,.png,.jpg,.jpeg} 
\else
  \usepackage{graphicx}                
  \DeclareGraphicsExtensions{.eps}     
\fi%

\graphicspath{{figures/}{pictures/}{images/}{./}} 

\usepackage{microtype}                 
\PassOptionsToPackage{warn}{textcomp}  
\usepackage{textcomp}                  
\usepackage{mathptmx}                  
\usepackage{times}                     
\usepackage{cite}                      
\usepackage{tabu}                      
\usepackage{booktabs}                  

\usepackage{array}
\newcommand{\etal}{\textit{et al}. }
\newcommand{\ie}{\textit{i}.\textit{e}. }
\newcommand{\eg}{\textit{e}.\textit{g}. }
\usepackage{balance}
\usepackage{siunitx} 
\usepackage{amsfonts} 
\usepackage{amsmath} 
\usepackage{multirow} 

\onlineid{2403}

\vgtccategory{Multi-sensory Experiences and Methodological Issues}

\vgtcinsertpkg

\title{Passing a Non-verbal Turing Test: \\Evaluating Gesture Animations Generated from Speech}

\author{Manuel Rebol\thanks{e-mail: rebol@student.tugraz.at}\\ %
    \scriptsize Graz University of Technology %
         \and Christian Gütl\thanks{e-mail: c.guetl@tugraz.at}\\ %
 \scriptsize Graz University of Technology
      \and Krzysztof Pietroszek\thanks{e-mail: pietrosz@american.edu}\\ %
     \scriptsize American University 
}

\teaser{
  \includegraphics[width=.5\textwidth]{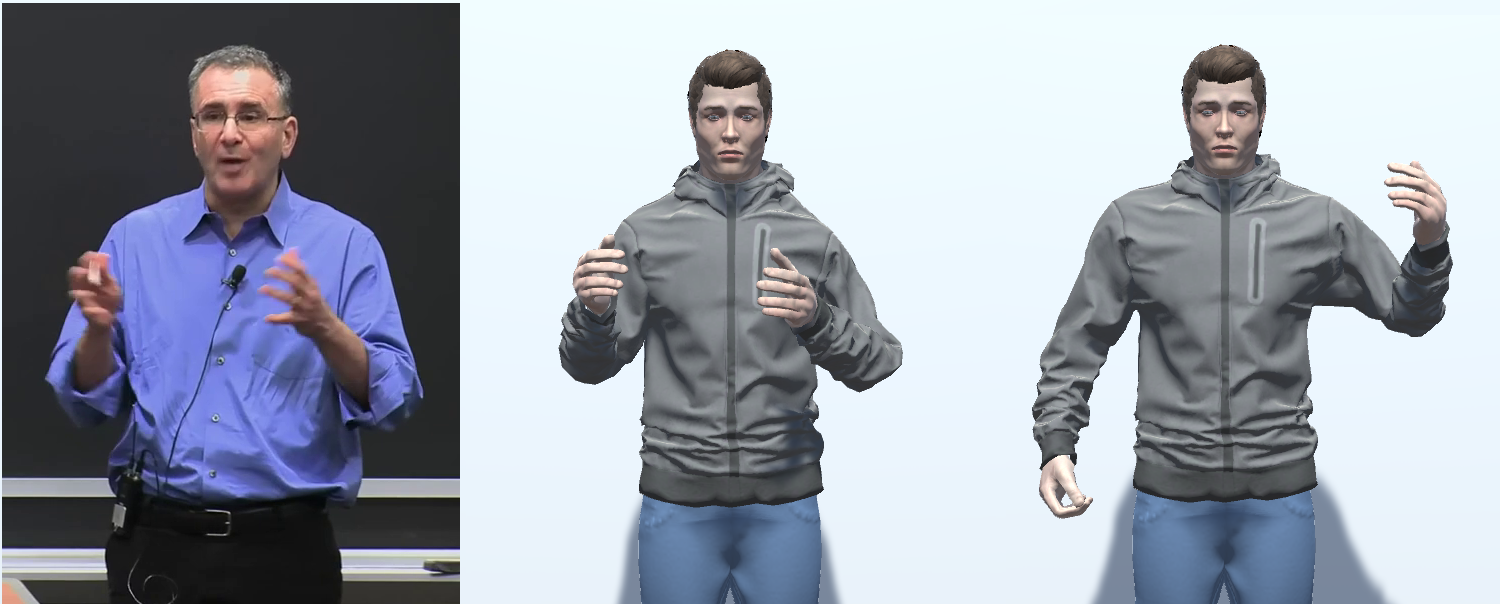}%
  \includegraphics[width=.5\textwidth]{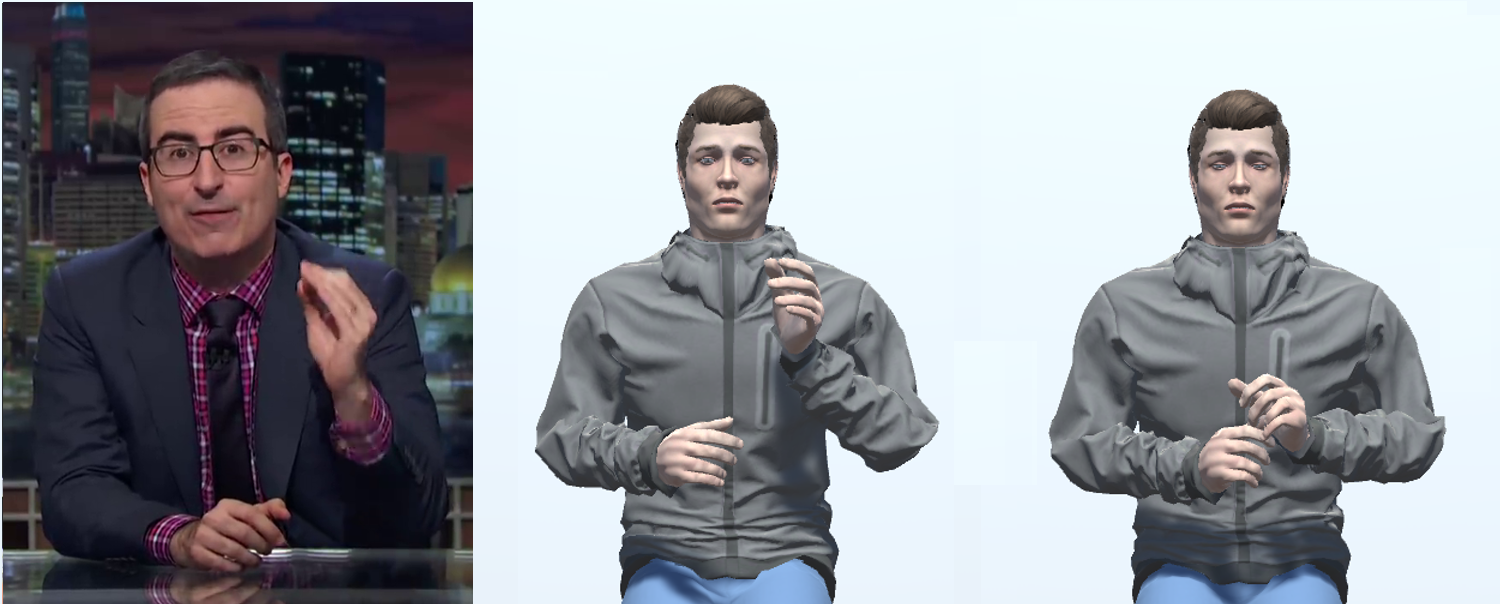}
  \caption{Gesture generation for professor Jonathan G. (left: original video frame, middle: ground truth, right: generated gesture animation) and television show host John O. (left: original video frame, middle: ground truth, right: generated gesture animation)}
  \label{fig:teaser}
}

\abstract{
People communicate using both speech and non-verbal signals such as gestures, facial expression, or body pose. Non-verbal signals impact the meaning of the spoken utterance in an abundance of ways. An absence of non-verbal signals impoverishes the process of communication. Yet, when users are represented as avatars, it is difficult to translate non-verbal signals along with the speech into the virtual world without specialized motion-capture hardware. 
In this paper, we propose a novel, data-driven technique for generating gestures directly from speech. Our approach is based on the application of Generative Adversarial Neural Networks (GANs) to model the correlation rather than causation between speech and gestures. This approach approximates neuroscience findings on how non-verbal communication and speech are correlated. 
We create a large dataset that consists of speech and corresponding gestures in a 3D human pose format from which our model learns the speaker-specific correlation.  
We evaluate the proposed technique in a user study that is inspired by the Turing test. For the study, we animate the generated gestures on a virtual character. 
We find that users are not able to distinguish between the generated and the recorded gestures.
Moreover, users are able to identify our synthesized gestures as related or not related to a given utterance.
Code and videos are available at \small \url{https://github.com/mrebol/Gestures-From-Speech}
} 


\CCScatlist{
\CCScatTwelve{Computer systems organization}{Neural networks}{}{}
\CCScatTwelve{Computing methodologies}{Activity recognition and understanding}{}{}
  \CCScatTwelve{Computing methodologies}{Animation}{}{}
  
}

\begin{document}

\maketitle

\section{Introduction}
Non-verbal communication is an essential component of the communication process, as it conveys a large portion of the message's meaning.
Thus, equipping digital representations of humans with non-verbal communication skills finds many applications in human-computer systems, from telehealth to entertainment to education. 

Despite ongoing research in human-machine communication, the interaction with virtual humans does not feel similar to talking to a person. 
We tackle the problem of unrealistic human-agent communication in virtual reality. 
We aim for synthesizing and animating \emph{natural} non-verbal communication in the form of hand and arm gestures. 
Our objective is to improve the interaction between humans and machines. Whenever people communicate, body language plays an important role. Hand and arm gestures are performed to support spoken language during communication and to convey meaning effectively. Moreover, speech and gestures reflect the personality and the intention of a person which is an important factor in virtual online communication. Gestures are idiosyncratic and contain information such as personal feelings, empathy, aggressivity, as well as authority and authenticity of a human \cite{KRAUSS1996389, gesturesEmotion2009}. The meaning of spoken language highly depends on body language because it has the power to signal irony and even reverse the sense of words pronounced. Therefore, the relationship between speech and gestures is of great importance when animating virtual humans. 

In this paper, we propose a novel, data-driven approach to generating meaningful gestures. 
We utilize the large amount of speaker video data available to train a model such that it learns the relationship by observing samples of speakers.
We adopt the state-of-the-art human pose estimation algorithms \cite{openpose1, videopose3d2019, hand3d2017} to extract the 2D pose from video and to project it into 3D space. Using our extraction framework, we create a 3D speech-gesture human pose dataset with two educational speakers and two comedians. Our resulting dataset is significantly larger than previous datasets \cite{JapaneseDataset, gestureRNN} that are recorded using motion capture. Moreover, our approach of creating 3D speech-gesture data can be applied to any speaker with exiting video material. In contrast to the dataset created by \cite{speech2gesture2019}, our dataset consists of human pose data in 3D space.
We design a Generative Adversarial Network (GAN) \cite{goodfellow2014generative} framework and apply deep learning to train a model that generates speaker-specific gestures from speech. We train our model on different speakers from our 3D speech-gesture dataset. Compared to the GAN framework of \cite{speech2gesture2019}, our architecture predicts 3D pose data. Moreover, we enforce constant bone length during the training of our GAN model to generate anatomically plausible gestures.  
Once we have generated gestures, we post-process the human pose data further such that we animate smooth and plausible gestures on avatars in a virtual environment. To achieve that, we apply skeletal constraints and inverse kinematic computations.
We design a user study to evaluate the qualitative results of our trained gesture model. 
In contrast to other recent approaches  \cite{speech2gesture2019, gesturesGAN, gestureGenerative} which perform the user study directly on the gesture output in human pose format, we animate the generated gestures on an avatar in a virtual environment. Hence, instead of comparing abstract skeletal representations of gestures, our study setting is natural and closely related to watching real humans communicate. Our user study is set up in a format that is inspired by the Turing test. We compare the gestures extracted from the original video to our generated gestures side-by-side and ask the participants of our study to identify the ``real'' gestures extracted from the video source. In a second experiment, we compare our generated gestures against gestures from an uncorrelated speech of the same speaker. 
We show examples of speakers from two different domains and their original and generated gestures animated on a virtual character in \autoref{fig:teaser}.

We discuss related work in the area of gesture synthesizing in the next section. Then, we describe our approach, consisting of the 3D gesture dataset creation, the gesture-speech translation with the GAN, and the animation on a virtual character, in detail. We present the evaluation of our method that was done by conducting a user study in \autoref{sec:eval}. 

\section{Prior Work}
Over the last three decades, many techniques were developed to generate gestures and body posture for a virtual human \cite{levine2009real,marsella2013virtual,lhommet2013gesture}. Typically, the animation generation is divided into two stages. First, the gesture and body pose to be generated is identified by analyzing a speech or text of the utterance. In the second stage, the gesture and body pose is generated and applied to the virtual human. 

\subsection{Communication Theory}
Previously developed techniques often do not apply the communication theories and ignore the neurophysiological basis of gestures. Our insight is supported by the Gestural Theory \cite{hewes1973primate} according to which non-verbal communication is not secondary to the speech's semantics or prosodics. 
Rather, speech and gestures are co-related and emerge together from the communicative intention of the thoughts forming in the speaker's mind. The Gestural Theory also poses that gesturing is not used only as a conscious way to illustrate speech, but rather is an integral part of the thought process that \emph{generates} the speech \cite{mcneill1992hand}. To further support that claim, various studies provided evidence that gestures are most effective when they are complementary rather than redundant with the semantics of the speech \cite{singer2005children}.  Other theories of communication, such as McNeill's Growth Point hypothesis also state that gestures and language emerge in a shared process from a communicative intent \cite{mcneill2008gesture}. In fact, new cognitive models explore the distribution of communicative content across output modalities rather than assuming speech-to-gesture causality \cite{kopp2013spreading}. 

An important related dimension of virtual human non-verbal communication that is not addressed by previous animation synthesis techniques is the viewer's perspective. From the perspective of the viewer, the process of understanding a gesture performed by others (e.g. virtual human) is a combination of interpretations of auditory and visual signals rather than being a purely visual phenomenon. This is because humans are \emph{simultaneously} verbally \emph{and} visually-oriented \cite{verbalVisual1992}. The human brain expects to process visible body movement when we hear spoken words. If the speaker is not visible, humans recreate the missing visuals in their imagination. If the speaker's body language does not meet the expectation or, worse, clashes with the semantics of the utterance, the process of communication is negatively affected.

Research in the neurology of speech suggests that the brain's auditory processing of speech is based on the gestural production of sounds and the original vocal energy as referenced from the body, rather than being based on the acoustical attributes of the sound itself \cite{deacon1998symbolic}. This is because our brains first evolved to understand visible gesture and then re-adapted for vocal ``gestures'', to finally learn to process spoken words. 
Based on these observations, we conclude that the traditional approach to the generation of gestures from speech does not take into account communication theories and neuroscience findings. This disconnect was previously acknowledged by Chiu \etal, who wrote: ``It is believed that in human communication, the brain is co-planning the gesture and the utterance, so approaches that do not use future information about the planned utterance may be unlikely to match the sophistication of human gesture-speech coordination.''\cite{chiu2015predicting}. 

\subsection{Gesture Generation}
An example of successful generation of beat gestures was presented by Levine \etal \cite{levine2009real}, who generated co-verbal gestures from speech by extracting \emph{prosodic} (tone and intonation) variations from the speech. The approach was unsuccessful in temporally matching the expressiveness of the speech with iconic or metaphoric gestures, as the information required to do so is not present in the prosodic properties of the speech. This concern was addressed by combining prosody with parsing of the spoken text \cite{marsella2013virtual}. Text can be analyzed for emotional content and rhetorical style, providing a rich -- although imposed by text semantics and thus lacking subtleties of subtext interpretation -- basis for iconic and metaphoric gesture synthesis. Finally, text-driven rule-based approaches \cite{lhommet2013gesture} create a sequence of gestures, but the rules must be coded manually. As a result, the gesture animations produced using rule-based systems tend to be time-consuming, repetitive, and lacking in variation.

Gesture synthesis, once a gesture to be generated is identified, often uses kinematic procedural techniques \cite{hartmann2005impl}. As an alternative, Kopp and Wachsmuth \cite{kopp2004synthesizing} presented a neurophysiologically-based approach to drive the trajectory of gesturing arm motions. Physics-based simulation has also been implemented, achieving better visual fidelity than kinematic-based approaches \cite{welbergen2010elckerlyc}. Procedural techniques enable full control of the motion, allowing to map the temporal dimension of the gesture to the temporal dimension of the speech. Unfortunately, they usually result in unnatural-looking hand movement and thus cause the uncanny valley effect \cite{seyama2007uncanny}. 

To mitigate the uncanny valley effect, techniques based on motion capture data have seen increased use. The use of motion graphs \cite{kovar2008motion} allows concatenated segments of motion captures to create a sequence, such as in \cite{muller2005efficient, pietroszek2017real}. Chi \etal \cite{chi2000emote} use the effort and shape components of Laban Movement Analysis to provide an expressive parameterization of motion. Hartmann \etal \cite{hartmann2005implementing} use tension, continuity, and bias splines to control arm trajectories and provide expressive control through parameters for activation, spatial and temporal extent, fluidity, and repetition. Pietroszek \etal proposed synthesizing motion by comparing rough input with a large number of high-quality motion templates in real-time using a compute shader implementation of Dynamic Time Warping \cite{pietroszek2017real}. Project SAIBA that combines the efforts of several research groups developed ``behavior realizers'' \cite{vilhjalmsson2007behavior}. The project is based on animation engines capable of realizing commands written in the Behavior Markup Language \cite{vilhjalmsson2007behavior}. These systems emphasize control and use a combination of procedural techniques and motion captures, but they do not allow the generation of gestures in real-time from speech.

Machine learning data-driven approaches were also applied to predict the gestures that may be specific to a given person's gesturing style \cite{kipp2005gesture}. Recent work proposed applying deep learning to the mapping from text and prosody to gesture \cite{chiu2015predicting}. Prior work \cite{kopp2004synthesizing} also addressed modeling individual gesture styles through analyzing the relationship between extracted utterance information and a person's gesture. While earlier works based on this approach have focused on addressing the mapping relation between linguistic features and gestures \cite{kita1990temporal}, recent work \cite{kopp2013spreading} has also addressed how to use acoustic features to help gesture determination. Chiu \etal \cite{chiu2015predicting} proposed a data-driven model that builds upon prior work \cite{artieres2010neural} by combining the advantages of deep neural networks for mapping complex relationships with an undirected second-order linear chain for modeling the temporal coordination of speech and gestures.

Recurrent neural networks (RNNs) such as long short-term memory (LSTM) \cite{hochreiter1997lstm} or gated recurrent units (GRU) \cite{cho2014properties} have been at the forefront of human motion prediction. However, such deep neural networks based models are primarily deterministic. There have been attempts to modify RNN encoder-decoder frameworks to work as a combination of deterministic and probabilistic human motion prediction models \cite{fragkiadaki2015recurrent}. In recent advances, Jain \etal \cite{jain2016structural} have introduced several structures and frameworks with deep RNNs to produce state-of-the-art results in human motion prediction. 
In contrast to RNN approaches, \cite{butepage2017deep} proposed a sparse autoencoder model to predict human motions in an unsupervised manner without recurrent units. Even though these models are good at human motion prediction, applying them to the generation of novel human actions has been a challenge.

Most recent approaches focus on the non-deterministic relationship between speech and gestures.
To overcome the problem of predicting the mean gesture, Alexanderson \etal \cite{gestureGenerative} introduce a probabilistic approach. They adapt the MoGlow framework \cite{henter2019moglow} which implements LSTM cells to generate gestures with realistic motion. Ferstl \etal \cite{gesturesGAN} use a GAN model in their non-deterministic approach. They train the adversary to generate realistic sub-features such as plausible dynamics and smooth motion. 
In contrast, Ginosar \etal \cite{speech2gesture2019} implement a single adversary which ensures realistic motion. Compared to previous approaches, they use in-the-wild YouTube video data to train their speaker models. It allows them to generate models for speakers from different genres, \ie show business, education and religion. Furthermore, their dataset is much larger than previous datasets. For a given set of speakers, they process more than 20 hours of data. This allows the models to learn from a variety of different input gestures. One drawback of the work of Ginosar \etal compared cannot be used in applications that require virtual humans. Although previous authors \cite{speech2gesture2019, gesturesGAN, gestureGenerative} evaluate the results comprehensively, they conduct the human study using gestures in the skeletal human pose format which is different from observing humans in reality. Thus, for the evaluation of our generative approach, we animate the gestures on virtual characters which are visually more closely related to real human communication. 

\section{Our Approach}

We model the correlation between body language and speech without imposing a causality between them. 
We consider learning models that correlate the input and output rather than create a causal relationship between them. As a result, rather than considering discriminative models that model the conditional probability $p(y|x)$, which can be interpreted as a causal relationship from $x$ to $y$, we focus only on generative models that model the joint probability distribution $p(x,y)$, which can be interpreted as a co-relational relationship between x and y. 

Translating our theoretical approach into the computation, we design a GAN architecture that predicts gestures given input speech. The goal of the network is to learn the co-relation, not causality, in a \emph{$<$speech, gesture$>$} pair. Once trained, the GAN generates the \emph{gesture} sequence given the \emph{speech} sequence. Our non-verbal communication animation model is therefore data-driven.
The GAN takes speech as input and predicts the gesture animation as output. 

\begin{figure*}
	\centering
	\includegraphics[width=\textwidth]{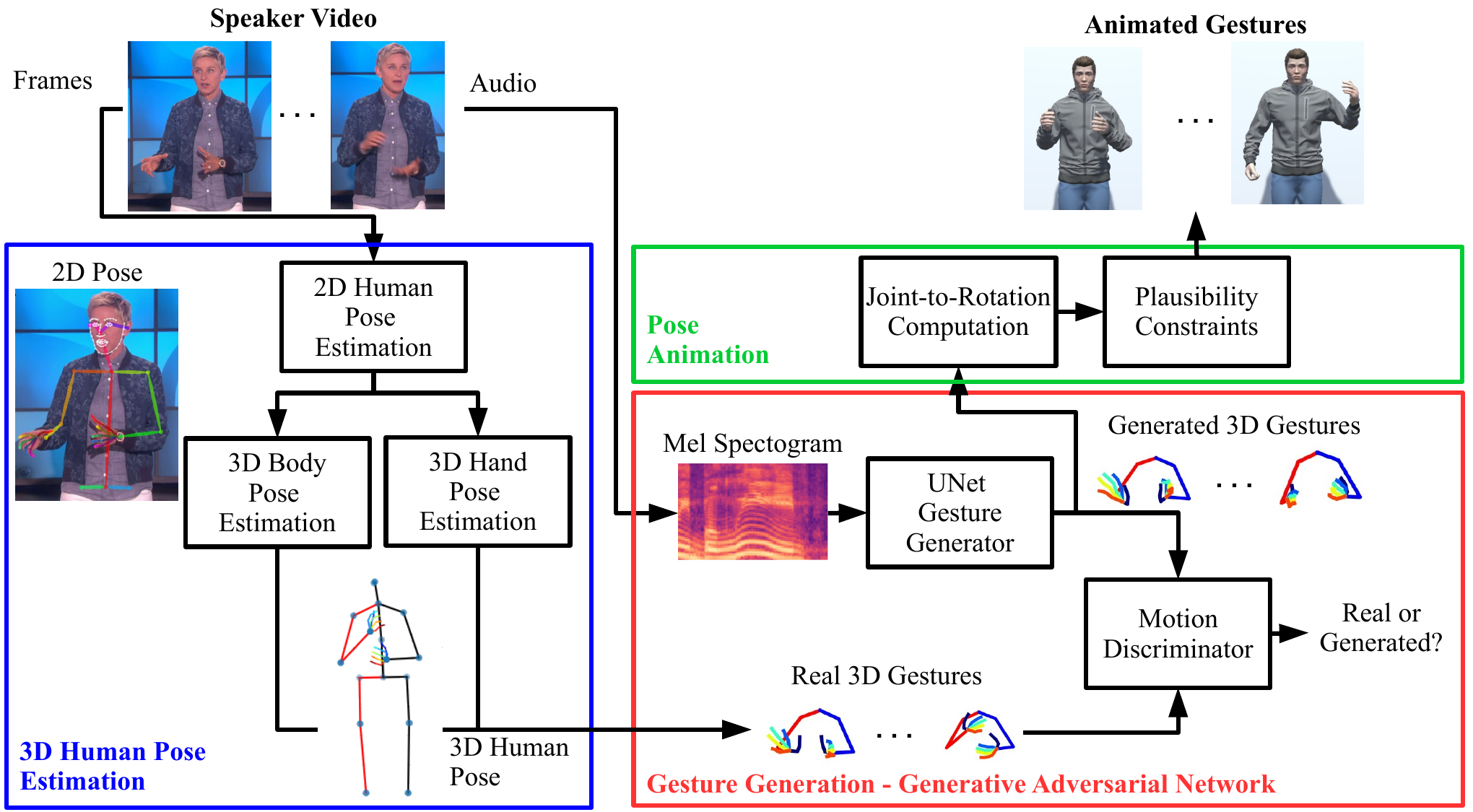}
	\caption{Gesture pipeline. Our gesture generation pipeline takes videos of speakers as an input (top left) and produces animated gestures (top right). In between, three major components exist: 3D Human Pose Estimation (blue), Gesture Generation with a Generative Adversarial Network (red) and Pose Animation (green). 
	\label{fig:gesture-pipeline}}
\end{figure*}

Our method is divided into three core components. We illustrate the structural overview in \autoref{fig:gesture-pipeline}. 
The gestures in human pose format are provided by our first component 3D Human Pose Estimation. We estimate the human pose in each frame of input video sequences. Once the 2D human pose is estimated, it is projected into 3D space and handed over to the GAN. 
The gesture generation GAN represents the main component. Our GAN takes two inputs during training, the raw audio of the speech and the gestures in human pose format. We extract the raw audio from our video dataset. 
The third component animates the generated gestures on a virtual character. This component computes the rotation angles of body parts given pose data and enforces physically plausible animations.

\subsection{3D Human Pose Estimation}
The amount and the quality of the training data determine the predictive power of our model. In the multimodal task of gesture generation, it is of utmost importance to have a large dataset that contains a wide variety of gestures. Unfortunately, existing motion capture datasets of gestures do not suffice, as they are neither large enough nor sufficiently labeled. Instead of relying on existing, insufficient datasets, we generate the training data from another readily-available large source of data: 2D videos of people performing lectures and speeches. Recent advancements in motion capture estimation from 2D video \cite{chen20163d,zhou2016sparseness,du2016marker} allow us to process many online videos of lectures and speeches and extract 3D body poses and gestures from them.

We extract the gestures from video data in the 3D human pose format. This efficient representation of gestures allows us to extract only the information required for gesture modeling. By extracting the human pose, we ignore image information such as background, color, and shape, which is irrelevant to our task.
We take the following precautions to ensure high quality of data, both before and after extracting body language from the video. In the process, we ensure that the extracted gestures are still temporally synchronized with audio. Before extracting motion from a video, we eliminate those fragments of the videos where the face, arms, or hands of the speaker are not entirely visible, \eg because it is being partially occluded by an object. 

We approach the gesture extraction in two steps. In the first step, we extract the 2D Human Pose from the raw video using the OpenPose framework \cite{openpose1, openpose2}. In the second step, we project the 2D pose into 3D space for the large body parts and hands separately. For the 3D body pose estimation, we use the model implemented by Pavllo \etal \cite{videopose3d2019}. For the 3D hand pose estimation, we implement a model similar to Zimmerman \etal \cite{hand3d2017}. We create a continuous heatmap by applying Gaussian kernels at the estimated OpenPose 2D keypoint positions. We also add temporal smoothing to improve video processing. 

As a result of the above process, we created a large dataset of motion captures that correspond to the gestures used by humans when speaking. Throughout the process, we preserved the time synchronization between the audio recording of the speech and the extracted motion corresponding to the speech. The extracted motion capture sequences serve as the $speech$ time series that is labeled by the $gesture$ time series during the supervised training of our network.

\subsection{Gesture Generation - Generative Adversarial Network}
We implement the Generative Adversarial Neural Network (GAN) \cite{goodfellow2014generative,mirza2014conditional,arjovsky2017wasserstein,zhang2017stackgan} framework that allows us to model the multimodal task of predicting gestures from speech. Inside the GAN framework, we locate the gesture generator $G$ and the motion discriminator $D$. 

\paragraph{Gesture Generator} The objective of the gesture generator is twofold. First, the generator is trained to predict gestures close to the ground truth gestures extracted from the input video. Second, the main objective of the generator inside the GAN framework is to fool the discriminator. Consequently, the predicted gestures should have realistic motion. 

We implement the UNet architecture introduced by \cite{unet} for our generator. The UNet models the relationship between the gestures encoded in human pose format and speech. Inside the UNet, the skip connections forward low-level prosodic features extracted from the input audio. These features are necessary to predict smaller beat gestures. 
The UNet bottleneck extracts high-level features that contain information about long input sequences. This is used to predict the posture of the speaker. 

We introduce the following loss function to train the parameters of our UNet model:
\begin{align} \label{eqn:gen}
\mathcal{L}_\text{Gen}(G) \ = \ &\mathbb{E}_{(\mathbf{s},\mathbf{p})}[ ||\mathbf{p} - G(\mathbf{s})||_1 ] \ + \\ &\lambda_\text{bone} \ \mathbb{E}_{(\mathbf{s})}[ ||B(G(\mathbf{\mathbf{s}}_t)) - B(G(\mathbf{s}_{t-1}))||_1 ]\notag ,
\end{align}
where vector $\mathbf{s}$ refers to the input speech and vector $\mathbf{p}$ refers to the pseudo ground truth body keypoints. The function $B$ returns the bone lengths in 3D space by computing the euclidean distance between pairs of keypoints at consecutive time steps $t$ and $t-1$.

The first term in \autoref{eqn:gen} ensures that model learns to predict pose positions $G(\mathbf{s})$ that are similar to the ground truth pose positions $\mathbf{p}$ extracted from the video.  
The second term \autoref{eqn:gen} ensures that the predicted bone lengths stay constant over time. This term forces the model to learn about the anatomical constraint of constant bone length in the human skeleton. The hyperparameter $\lambda_\text{bone} \in (0,1)$ is used to weight the importance of constant bone length term with respect to the other terms within our objective function defined in \autoref{eqn:Whole}. 
During training, we accept a higher loss from the first term to avoid regressing to the mean pose. The loss of the second term remains low.

\paragraph{Motion Discriminator} 
The communication theory suggests that the same utterance can lead to different gestures. Hence, the regression loss between generated gestures and pseudo ground truth does not model the relation between speech and gestures as a whole. The problem is the regression towards the mean pose. For example, if the input data includes the same utterance twice where one time the pose moves into the direct opposite direction of the other time, the model learns to predict no motion. When considering the whole training dataset, this will result in predictions with less motion compared to the training dataset. To avoid this phenomenon, our discriminator inside the Gesture GAN architecture ensures that the motion of the generated gestures is similar to the motion extracted from video. 

The discriminator receives a motion sequence, either predicted by the generator or obtained from the input video, as input. The motion is computed by taking the difference between consecutive pose vectors. We express this by introducing the function $M$
\begin{equation} 
\label{eqn:motion}
M(\mathbf{v}) = \mathbf{v}_{t} - \mathbf{v}_{t-1} \ ,
\end{equation}
which computes the motion between two consecutive pose vectors $\mathbf{v}_{t}$ and $\mathbf{v}_{t-1}$. 

The objective of the discriminator is to detect if a given gesture sequence is real or generated, while the generator tries to fool the discriminator. Consequently, we have created a two-player optimization game. As a result, the generator tries to predict gestures similar to the real gestures in terms of the motion. The discriminator tries to identify the predicted gestures as such, even as they get more realistic during training.

The complete GAN loss function including the discriminator $D$ is defined as 
\begin{equation} 
\label{eqn:gan}
\mathcal{L}_\text{GAN}(G, D) = \mathbb{E}_{(\mathbf{p})}[\log D(M(\mathbf{p})) ]+ \mathbb{E}_{(\mathbf{s})}[\log(1 - D(M(G(\mathbf{s}))))] \ .
\end{equation}
The objective of the discriminator is to maximize this function. Consequently, the term loss is only true concerning the generator. The discriminator is trained to output $D(\cdot) \rightarrow 1$ if input motion is real and $D(\cdot) \rightarrow 0$ if the input motion is generated.

\paragraph{GAN Objective}
We train the parameters of our model by combining the loss functions shown in \autoref{eqn:gen} and \autoref{eqn:gan}. 
The final objective function is defined as
\begin{equation} \label{eqn:Whole}
\min_G \max_D \mathcal{L}_\text{GAN}(G, D) + \mathcal{L}_\text{Gen}(G)	.
\end{equation}
The generator $G$ has the objective to minimize this function whereas the discriminator $D$ aims to maximize $\mathcal{L}_\text{GAN}$.

\subsection{Pose Animation} 
Our Gesture GAN model produces 3D Human Pose sequences which we animate on virtual humans using rotation angles between bones and inverse kinematic computations. By connecting the keypoints predicted by our Gesture GAN, we create a skeletal representation of the human pose. We use this skeletal representation of the gestures to animate an avatar in a virtual environment.

One challenge when transferring the predicted 3D Human Pose into a 3D animation is the missing information about anatomical details and ambiguities. Since we animate the gestures on a virtual human of different size and shape compared to the original speaker, we omit the information about the bone length of the generated skeleton. Instead, we only consider the rotation between the bones. 
When viewing the angles between two bones in the Euler angle representation, we obtain the pitch and yaw angles from the skeleton. However, we do not have any information about the roll angle. To tackle this problem, we approximate the roll rotation of different body parts using inverse kinematics.

Besides recovering the missing information about the roll angle, we also have to deal with the second problem introduced when converting the human pose representation to a virtual human animation, which is implausible motion. Although our Gesture GAN is trained to predict motion which is similar to real human motion, there exists no constraint which particularly enforces anatomically plausible motion. Hence, the predicted motion in some cases appears to be very artificial, especially when animated on a virtual human avatar. This problem is very distracting for the viewer of the animation because anatomically implausible motion is quickly noticed. To overcome this issue, we introduce motion constraints on the fingers. The motion constraints enforce that the fingers can only be bent in anatomically possible angles and directions. In addition to the motion constraints, we apply motion smoothing. 

\subsection{Training the Network} 
For the purpose of training our GAN, we generate a gesture dataset of over one hundred hours extracted from videos of four speakers from two domains. Specifically, we generated 72 hours of motion capture data for television show hosts John O. and Ellen D. as well as 64 hours of motion capture data for professor Jonathan G. and Shelly K. We encode the gestures in the efficient human pose format. By using this format, we ignore irrelevant information such as background and the shape of different body parts of the speaker. 

We feed the extracted gestures in 3D human pose format to the discriminator inside our GAN architecture. 
The input to our GAN generator is the Mel spectrogram of the audio directly extracted from the training videos of each speaker. We sample the audio data at a rate of 16,000 samples per second. In order to provide the generator with the temporal context, the input interval is four seconds long, which was found to be the best length empirically.
We evaluate the predictions of our Gesture GAN on our validation set using the Percent of Correct Keypoints (PCK) \cite{pck2013} metric with proximity radius $\alpha = 0.2$. The quantitative results are shown in \autoref{tab:quantitative-results}. We observe that John O. and Shelly K. who are in sitting position and therefore show less upper body movement achieve a higher PCK. In contrast, the PCK is lower for speakers that are standing (Ellen D.) and walking around (Jonathan G.). 
We select the speakers John O. because of the high PCK and Jonathan G. because we want to examine both sitting and standing speakers for further experiments in our user study.

\begin{table}
	\renewcommand{\arraystretch}{1.2}
	\setlength{\tabcolsep}{0.6cm}
	\centering
	\begin{tabular}{l|l|l}
		 \textbf{Category} & \textbf{Speaker} & \textbf{PCK $\uparrow$} \\ \hline  \hline
		 \multirow{2}{*}{TV Show}& Ellen D. & 31.0 \\  
		 & John O. &  \textbf{60.3} \\ \hline  
		 \multirow{2}{*}{Education}& Shelly K. & 40.6 \\  
		 & Jonathan G.& 23.6 \\  
		\hline
	\end{tabular}
	\vspace{0.1cm}
	\caption{Quantitative evaluation of gestures. We evaluate the four speakers from the two categories TV Show and Education on our validation datasets. For comparison we use the Percent of Correct Keypoints (PCK) metric. 
	}
	\label{tab:quantitative-results}
\end{table}

\section{Experimental Evaluation} \label{sec:eval}
In order to validate the perceptual quality of the synthesised animations, we performed a user study designed to answer the following questions:
\begin{enumerate}
    \item Can users detect difference between the synthesized gesture and the ground truth (original) gesture for a given speech fragment?
    \item Are synthesized co-verbal gestures perceived by users as correlated with a given speech fragment?
\end{enumerate}
To answer the above questions, we designed two experimental tasks.

\begin{figure} 
	\centering
	\includegraphics[trim={80 0 0 0},clip=true, width=1.0\columnwidth]{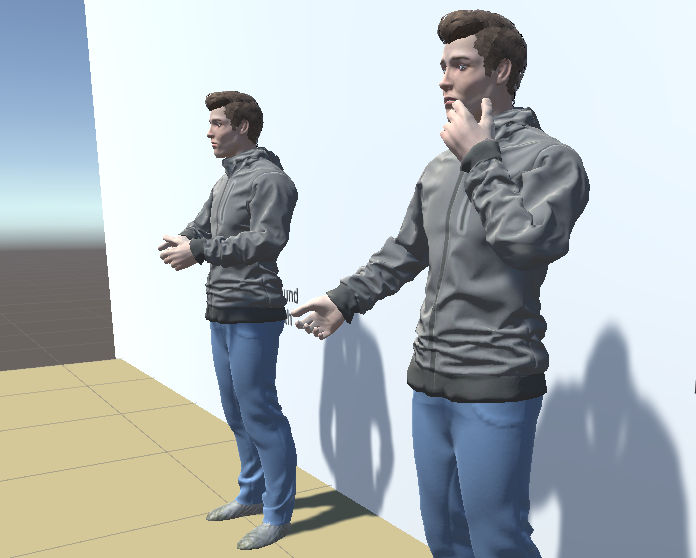}  
	\caption{Virtual scene. The scene shows the two virtual characters performing non-verbal communication next to each other. We use this setting for our side-by-side comparisons.}
	\label{fig:3dscene}
\end{figure}

In the first task, we present users with a series of side by side videos of two identical avatars performing different gestures for the same, 12-second long speech fragment that can be heard by the users. One of the videos shows the ground through gestures, as extracted from the video recording of the speaker and applied to the avatar. The other video shows the gestures synthesized by our system from the same speech fragment. The left-right position of the videos is randomized. The user is asked to decide which sequence of gestures is the ground truth gesture for the speech fragment.

In the second task, we also present the user with a series of videos of two identical avatars performing different gestures for 12-second long speech fragment. In this tasks gesture sequences for both avatars are synthesized. However, one of the gestures sequence, either right or left, is synthesized from the speech fragment heard by the participant, while the other avatar is performing gestures synthesized for a different, randomly selected speech fragment from the same speaker. The user is asked to decide which sequence of gestures is the ground truth gesture for the presented speech fragment.

In both tasks, the user can hear the speech and see the avatar performing the gestures, but the face of the gesturing avatar is hidden. We consider face expression and lips synchronization a confounding factor in our study, resulting from the uncanny valley effect. 
Our experimental design is inspired by the original Turing test \cite{turing2004computing}. 
Our system ``passed'' our test, if after a limited number of attempts, the human judges are unable to establish a statistically significant difference between the sequence of gestures generated by our system and the real gestural sequence.

\subsection{Implementation and Apparatus}

For the visualization of our generated gestures, we animate the UMA 2 Multipurpose Avatars using the Unity 3D game engine version 2019.3. Both animations of the side-by-side comparison run on two identical avatars, see \autoref{fig:3dscene}. The only difference between the avatars are the gestures that they perform. The avatars are placed against a neutral background, with a light source in front of them. The avatars are positioned at a distance of 1.5 meters next to each other, such that they do not occlude each other while gesturing. Both avatars face the camera at the same \ang{180} angle. The camera is positioned three meters in front of the avatars. The camera angle and shot size is adjusted such that it records the upper body of both avatars without showing the face.

\begin{figure}
    \centering
    \includegraphics[width=.98\columnwidth]{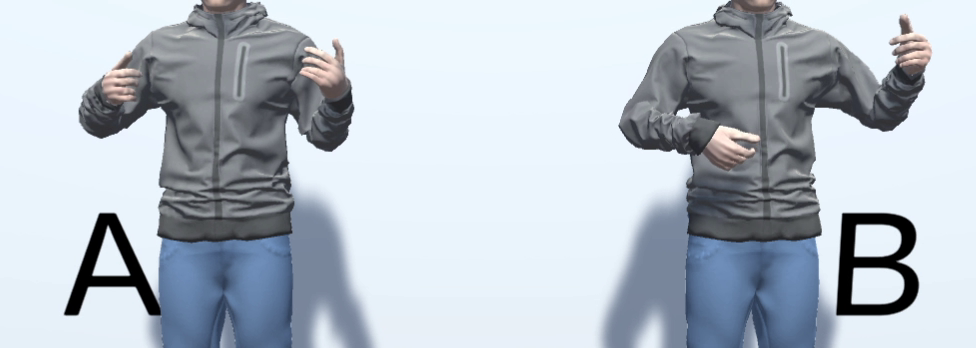}
    \caption{User study. The participants in our study were asked to decide which animated gestures they perceived as more natural. The gestures were animated on two identically looking avatars, referred to as A and B.}
    \label{fig:user-study}
\end{figure}

While the avatars are performing the gestures in 3D space, the virtual camera records the animation in 2D at the resolution of $1000\times360$ at 30 frames per second. We record the avatars from the hip upwards to only capture the information relevant for gesture evaluation and to eliminate the face expression as a confounding factor. An example of our setting is shown in Figure \ref{fig:user-study}.

\subsection{Null Hypotheses}
For the first task, the null hypothesis is that the proposed gesture synthesis is perceptually indistinguishable from the ground truth. 
For the second task, the null hypothesis is that the speech fragment is not correlated with the synthesised gesture.

\subsection{Participants}
For our study, we recruited 100 participants on Amazon Mechanical Turk (Mturk) platform. Each participant received a compensation of \$$3$. The requirement was that the participants are fluent in English. All participants finished the study. We did not collect any demographic data, or any identifiable data or information about participants' prior computer-related experience -- all collected data was anonymous at all time.  

\subsection{Procedure}
The participants were first presented with the description of the study and asked to confirm their willingness to participate. Then, the participants were presented with three repetitions of task 1 and task 2 as a training. Next, the participants were presented with a series of twenty 12-second long videos for task 1 followed by a series of ten 12-seconds long videos for task 2. In each task, half of the videos were generated for the speech of John O., while the other half were generated from the speech of professor Jonathan G.

For each task, the participants were asked to select the video that feels as a more natural representation of gestures for the presented speech fragment. Participants were forced to select one option, before they could proceed to the next video. Also, we ensured that it was not possible to select the answer without watching the whole video. Additionally, it was not possible to watch the same video twice. Once completed, the participants were presented with a ``thank you'' page.

\subsection {Results}
\begin{figure}
	\centering
	\includegraphics[width=\columnwidth]{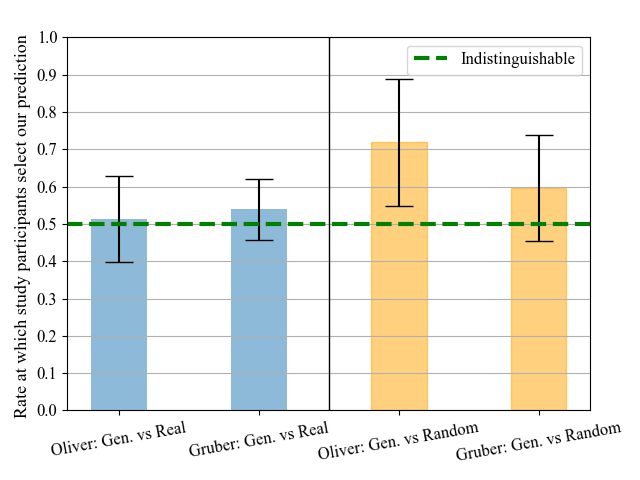}
	\caption{Quantitative results. The four bars refer to the four different tasks the participants completed during the user study for the speakers John O. and Jonathan G. The height of the bar represents the rate at which the participants selected the gestures predicted by our model. The vertical intervals at the top of each bar indicate the standard deviation across the participants.}
	\label{fig:bar-chart}
\end{figure}

We illustrate the quantitative results for both speakers in \autoref{fig:bar-chart}. We separate the two tasks for each speaker. 

For task 1, determining which gestures are generated and which are ground truth, the participants mistook synthesized gestures as ground truth $52.5\%$ of the time on average ($\sigma^2 = 1.0\%$). The difference between selecting ground truth and synthesized gestures were not significant. The results for individual speakers were not significant either, with participants selecting synthesized gesture over the ground truth, in average, $51.2\%$ of the time for John O. ($\sigma^2 = 1.4\%$), and $53.8\%$ of the time for professor Jonathan G. ($\sigma^2 = 0.6\%$). 

\begin{figure*}[ht]
	\newcommand{\cropimgGt}[1]{\includegraphics[ width=.25\textwidth]{fig/gruber/#1-crop}} 
	\newcommand{\cropimg}[1]{\includegraphics[ width=.25\textwidth]{fig/gruber/#1cut}} 
	\centering
	\cropimg{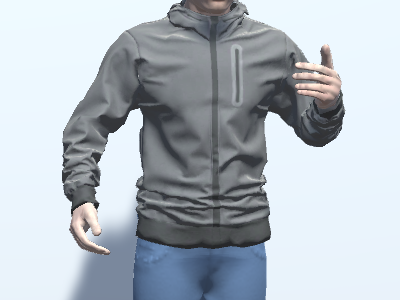}\cropimg{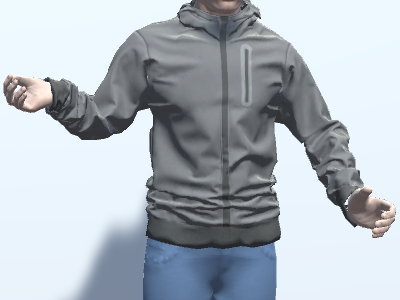}\cropimg{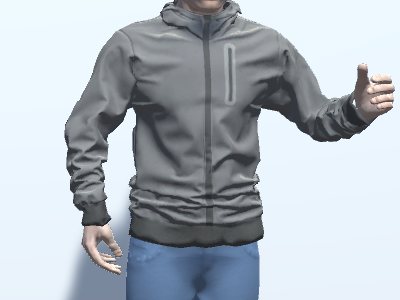}\cropimg{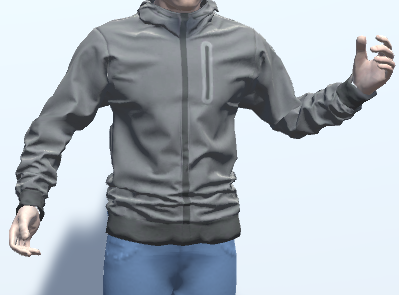}\\
	\cropimgGt{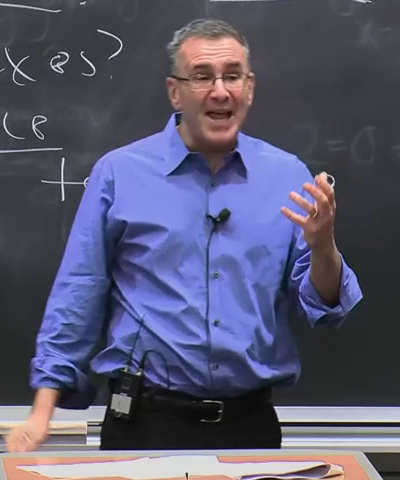}\cropimgGt{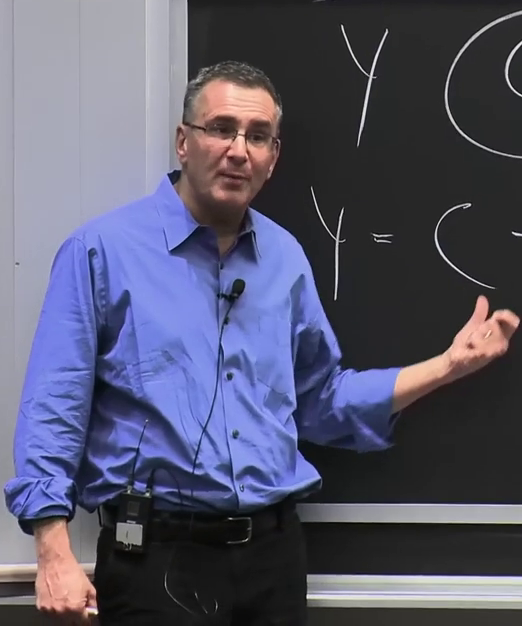}\cropimgGt{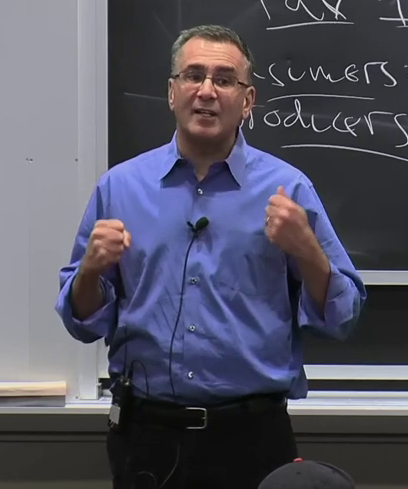}\cropimgGt{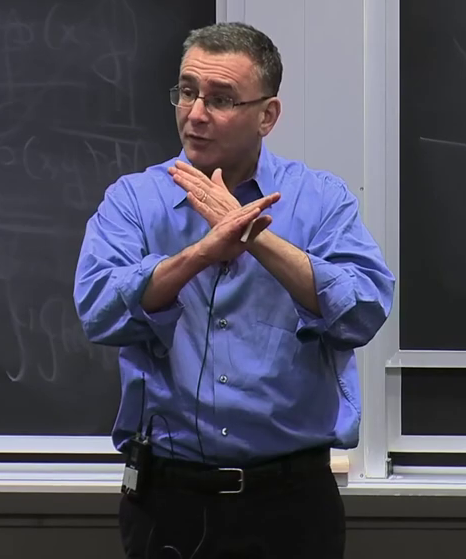}
	\caption{Qualitative results. We compare the predicted gestures animated on a virtual character (top) and the original video (bottom) of speaker 
	Jonathan G.
	Overall, the motion and the beat gestures are very similar. Whenever 
	Jonathan G.
	lifts his hand, the same motion is predicted by our model. In column two and three we show that sometimes only one or the other hand is lifted in the prediction. In column four, we show that metaphoric gesture ``intersection'' 
	is not predicted by our model. \label{fig:quality}}
\end{figure*}

For task 2, determining which gesture was generated for a given speech fragment, the participants chose the gesture sequence that was indeed generated for the given speech fragment in average $65\%$ of the time ($\sigma^2=3.55\%$). Interestingly, the participants were significantly more likely to point at a correct gesture sequence in case of gestures generated for John O., selecting it correctly $71.2\%$ of the time, while correctly identifying the gesture sequence, on average, $58.8\%$ of the time for gestures generated for professor Jonathan G. In both cases considered individually, the experiment showed significant difference between the gestures generated for the heard fragment of speech and gestures generated for some other fragment of speech from the same speaker.

\subsection{Discussion}
Based on the above results, we conclude that our system generates realistic body language, since the participants were unable to determine whether the gesture animations were generated or captured. It is important to note that our study does not attempt to establish that our gesture synthesis creates gesture sequences that are perceptually superior to the ground truth. Instead, we attempt to establish that, despite a relatively large and diverse sample, the participants are unable to reject the null hypothesis for task 1, while at the same time being able to reject the null hypothesis for task 2. In other words, the closer our system is to ground truth, the more difficult it is to reject the null hypothesis for task 1.

Rejecting the null hypothesis in task 1 would mean that our system generates gestures that are perceptually inferior as compared to the ground truth. Not rejecting the hypothesis, however, does not mean we have proven the equivalence between synthesized and extracted gestures -- it is not possible to do so. However, not rejecting the null hypothesis shows that a difference, if any, between the synthesized gestures and ground truth gestures must be small. 

The inability to reject the null hypothesis for task 1 could be a result of either the difference between the two groups being too small for our sample size to be significant \emph{or} it could be a result of the participants not paying attention to the study and selecting right or left videos at random. However, the rejection of the null hypothesis for task 2 serves as a validation for the claim that our generated gestures are indistinguishable from the real gestures, because finding significant difference in task 2 suggests,
that the participants were indeed paying attention to the presented videos and were not selecting answers A or B at random. 

We also took a closer look at each specific user's decision in judging the generated vs ground truth gesture animations. The analysis revealed that participants were fooled by our gesture prediction in those gesture sequences, in which the ground truth does not include an iconic gestures. Whenever the speaker performs an iconic gesture, subjects were able to identify the ground truth. This is an interesting result, since it suggests that the fact that our system does not model the speech semantics and therefore does not explicitly generate iconic gestures might provide our users with a winning strategy: searching for iconic gestures in the speech and identifying those. It remains an open question whether or not sufficiently large training dataset would result in the system encoding correlation between semantics of speech and iconic gestures. We give an insight into the qualitative results of speaker Jonathan G. 
in \autoref{fig:quality}.

Another qualitative observation is the difference in the correct recognition of gestures generated for a given speech fragment between John O. and Jonathan G. gestures. We observe that ground truth gestures of John O., a comedian, are more pronounced and exaggerated. Possibly, the gestures performed by John O., an experienced actor, may be better correlated with his speech than are the gestures of professor Jonathan G. If that is true, it may be possible to create a model that actually generates better gestures for a given speaker than the speaker's own gestures. 

One aspect that needs to be considered when interpreting our results is that we animated real and generated gestures on virtual humans which cannot be directly compared to a real human communication which is more complex. In real human communication, the human brain uses biological motion detection when watching living beings \cite{bioMotion2013}. Biological and nonbiological motion is detected differently as shown by Hiris \cite{hiris2007detection}. As a consequence, watching gestures of a speaker on video and animated gestures on avatars are judged differently by humans.

\section{Conclusion}
Generation of nonverbal communication, including gestures, is a missing cornerstone in creating virtual humans that communicate as efficiently as humans. Human-agent communication is used broadly in online learning environments, virtual assistance, and communication tools.

In this work, we predict the gestures from speech and animate them on virtual characters in a 3D environment. We implemented a novel 3D human pose estimation pipeline, which allows us to train speaker-specific gesture generation models from in-the-wild video data. Utilizing large-scale video datasets allows our model to learn from a variety of different gestures. 
We implement the non-deterministic relationship between speech and gestures using a GAN model.  

The results of our user study show that the gestures predicted by our model are natural and correspond to speech. 51.2\% and 53.8\% of participants selected that our generated gestures seem more natural than the original gestures for speakers John O. and Jonathan G., respectively. Furthermore, when comparing speech correlated and uncorrelated gestures from the same speaker, participants identified the speech correlated gestured correctly in 71.2\% of times for speaker John O. and 58.8\% for speaker Jonathan G. The results of both experiments indicate that our method can be applied successfully to speakers from different genres.

In future work, we will develop a model that synthesizes gesture animations given a speech stream of an arbitrary speaker. Therefore, data from multiple speakers needs to be combined such that a universal speech-gesture model can learn about how gesturing styles depend on the voice and prosody of humans.

\acknowledgments{
We want to thank Graz University of Technology and American University for maintaining a successful collaboration and supporting research stays abroad. 
This collaboration was endorsed by the Marshall Plan Foundation.}

\bibliographystyle{abbrv-doi}

\bibliography{gestures}
\end{document}